\documentclass[a4paper,10pt,twoside]{article}

\usepackage{url}
\usepackage{hyperref}
\usepackage{clin}        
\usepackage{harvard}     
\usepackage{xcolor}

\newcommand{\revision}[1]{\textcolor{blue}{#1}}

\begin{document}

\title{Analyzing Cancer Patients' Experiences with Embedding-based Topic Modeling and LLMs}

\author{Teodor-Călin Ionescu$^*$ \\
{\normalsize \bf Lifeng Han}$^*\dagger$ \email{l.han@lumc.nl}\\
{\normalsize \bf Jan Heijdra Suasnabar}$\dagger$ \email{j.m.heijdra\_suasnabar@lumc.nl}\\
{\normalsize \bf Anne Stiggelbout}$\dagger$ \email{a.m.stiggelbout@lumc.nl}\\
{\normalsize \bf Suzan Verberne}$^*$ \email{s.verberne@liacs.leidenuniv.nl}\\
\AND \addr{$^*$Leiden Institute of Advanced Computer Science (LIACS), Leiden, The Netherlands}
\AND \addr{$^{\dagger}$Leiden University Medical Center (LUMC), Leiden, The Netherlands} 
\\ Corresponding authors: LH and SV}

\maketitle\thispagestyle{empty} 


\begin{abstract}
This study investigates the use of neural topic modeling and LLMs to uncover meaningful themes from patient \revision{interview} data, 
with the goal of offering insights that could contribute to more patient-oriented healthcare practices. 
We analyze a collection of \revision{manually} transcribed interviews with \revision{pancreatic} cancer patients (132,722 words in 13 interviews). 
We first evaluate BERTopic and Top2Vec for the purpose of individual interview summarization, by using similar preprocessing, chunking, and clustering configurations to ensure a fair comparison on Keyword Extraction. LLMs (GPT4) are then used for next step topic labeling. Their outputs for a single interview (I0) are rated through a small-scale human evaluation, focusing on \textit{coherence}, \textit{clarity}, and \textit{relevance}. 
Based on the preliminary results and evaluation, BERTopic shows stronger performance and is selected for further experimentation using three \textit{clinically oriented embedding} models. 
We then analyzed the full interview collection with the best model setting.
Results show that domain-specific embeddings improved topic \textit{precision} and \textit{interpretability}, with BioClinicalBERT producing the most consistent results across transcripts. The global analysis of the full dataset of 13 interviews, using the BioClinicalBERT embedding model, reveals the most dominant \textit{topics} throughout all 13 interviews, namely ``Coordination and Communication in Cancer Care Management" and ``Patient Decision-Making in Cancer Treatment Journey''.
Although the interviews are machine translations from Dutch to English, and clinical professionals are not involved in this evaluation, the findings suggest that neural topic modeling, particularly BERTopic, can help provide useful \textit{feedback} to clinicians from patient interviews. This pipeline could support more efficient document navigation and strengthen the role of patients' voices in healthcare workflows.
Affiliated resources created in this work will be shared publicly at \url{https://github.com/4dpicture/TM4health} including codes on preprocessing and stopword list prepared. 
\end{abstract}

\section{Introduction}

Cancer is one of the most challenging global health issues, affecting not only individuals but also entire families and communities. People are subjected to intense physical and mental difficulties to the point where their quality of life changes forever, even after recovering from the disease. In modern healthcare, understanding patient experiences is crucial for improving treatment and care \cite{ren2025malei}. While clinical research traditionally relies on structured medical data, patient feedback, such as \revision{interview} data, provides valuable insights that should not be overlooked. Healthcare should not only focus on treating the disease, but also on the emotional and psychological needs of patients, recognizing them as individuals in need of comfort and support, rather than just subjects in a medical process. 
These have been reflected by recent work on shared decision making (SDM) and patient-centered carepath design \cite{griffioen2021bigger,kidanemariam2024patient,bak2025ethical}.

Natural Language Processing (NLP) is a powerful approach to analyzing patient narratives, enabling the extraction of meaningful topics from a large volume of text. In this paper, we use neural topic modelling techniques to extract relevant topics from cancer patient \revision{interview} data and analyze them systematically. \revision{Note that with the use of topic modeling on the interview collection, we do not aim to answer a clinician's information need; instead, we are trying to discover general patterns about patient experiences. This way,} we aim to see what kind of valuable insights we can offer to healthcare providers in general in order to make patients' cancer treatment journeys more bearable. \revision{The interview data is a valuable collection that is unique in depth: as opposed to experiences shared in online patient communities, the interviews are in-depth communication with a small group of patients over an extended period of time. Analyzing these data} could give professionals a deeper understanding of patient needs, enabling a more patient-oriented care strategy, \revision{beyond the individual patients included in the interview data}. We compare two topic modeling algorithms, BERTopic \cite{2022bertopic} and Top2Vec \cite{angelov2020top2vec,angelov-inkpen-2024-topic}, to determine which one performs better at extracting relevant topics from patient \revision{interview} data. By optimizing these models, we determine which topic modeling approach yields more interpretable and coherent topics within the context of \textit{cancer care}. Ultimately, this work lays the foundation for a potential feedback tool that allows clinicians to automatically analyze patient files, scanning through large bodies of text easily and focusing on key themes and concerns patients raise. For instance, for a collection of patient consultation transcripts, this methods can help to identify cross-patient population level concerns effectively.

We address two research questions in this paper:
\begin{enumerate}
    \item \textbf{\revision{Which topic modeling approach is the best choice for this task?} 
    }
    \item \textbf{\revision{What key themes can current neural topic modeling models extract from patient \revision{interview} data?} 
    }
\end{enumerate}

This study aims to connect patient feedback with clinical decision-making by addressing these questions. 

\revision{The dataset used in this study originates from a Dutch Cancer Society (DCS) Project of the Leiden University Medical Center (LUMC) \cite{griffioen2017potential,griffioen2021bigger,tudelft_metro_mapping}. The patients in the dataset were patients from the Erasmus Medical Center (Erasmus MC) in Rotterdam, and the data was provided by Erasmus MC}. \revision{The research conducted for this paper took place in the context of the 4D PICTURE consortium, an EU research project across multiple institutes from several countries\footnote{\url{https://4dpicture.eu}  ``Design-based Data-Driven Decision-support Tools: Producing Improved Cancer Outcomes Through User-Centred Research''}, which aims to improve the cancer patient journey and ensure personal preferences are respected.}By analyzing the cancer patient \revision{interview} data, we aim to contribute to that mission, proving insights that reflect the patients' lived experiences and can be used to further improve patient-centered healthcare practices.


\section{Background and Related Work} 

\subsection{Model Evolvement of Topic modeling}

Topic modeling is an unsupervised technique for identifying themes in large text collections and forms the foundation of this research. A widely used traditional method is Latent Dirichlet Allocation (LDA) \cite{LDA}, a Bayesian algorithm for extracting topics from text. LDA has been applied in various contexts, such as analyzing topic evolution during the COVID-19 pandemic in Swedish newspapers \cite{10337151}. However, LDA suffers from limitations, including difficulty capturing semantic relationships, sensitivity to common words, and issues with short-text data \cite{albalawi2020shorttext}. These shortcomings include non-deterministic outputs, the need for a predefined number of topics, data sparsity, and inability to model topic relations.

More recent research introduced embedding-based models like \textbf{BERTopic} \cite{2022bertopic} and \textbf{Top2Vec} \cite{angelov2020top2vec} include the contextual-embedding variation from \citeasnoun{angelov-inkpen-2024-topic}. \revision{Both models use pretrained language models to capture semantic similarity between documents. BERTopic builds on transformer-based sentence embeddings combined with HDBSCAN clustering, while Top2Vec jointly learns document and word embeddings in a shared vector space, enabling topics to emerge directly from dense semantic neighborhoods rather than predefined token distributions.  This overcomes} many LDA limitations \revision{that rely on literal terms}. \revision{In particular, Top2Vec works by first training or adopting a universal sentence embedding model to represent each document as a dense vector, then applying dimensionality reduction (typically UMAP) to preserve semantic structure in a lower-dimensional space. It then uses HDBSCAN to discover dense regions corresponding to latent topics, automatically identifying both the number and composition of these clusters. The model finally derives topic representations by identifying the word vectors closest to the centroid of each cluster, allowing for highly coherent topic descriptors that capture nuanced semantic relationships.} Both models require minimal preprocessing, automatically determine topic numbers, and excel with short or context-rich texts. \revision{This makes them especially effective for patient narratives or other domains where texts are informal, heterogeneous, or rich in contextual cues that traditional probabilistic assumptions fail to model. BERTopic uses a cluster-based TF–IDF to generate interpretable topic labels, and Top2Vec allows topics to be explored interactively by identifying documents similar to a topic based on cosine similarity.} Studies show \revision{that neural topic models} outperform classical methods in coherence and interpretability: 
\citeasnoun{egger2022interpretable} found that BERTopic produced the most distinct topics on COVID-19 Twitter data, while Top2Vec also surpassed LDA. Similarly, 
\citeasnoun{EvaluatingTopicModels2025} reported that BERTopic achieved the highest coherence scores across datasets, with Top2Vec ranked slightly lower but still ahead of LDA. Overall, embedding-based models consistently deliver more coherent and interpretable topics than traditional approaches.

\subsection{Patient-Narrative Clinical NLP} 

Clinical Natural Language Processing (NLP) applies NLP techniques, such as topic modeling, to extract and analyze medical text from sources like unstructured health data, discharge summaries, surveys, and patient files. Modern research highlights its integration into healthcare systems, particularly for \textit{clinical decision support} and \textit{clinician-patient communications} \cite{DEMNERFUSHMAN2009760,wang2018clinicalIEreview}. For instance, in the statistical era, the review by ~\citeasnoun{DEMNERFUSHMAN2009760} documents the use of the Linguistic Inquiry and Word Count (LIWC) tool for analyzing patient personality through linguistic style, enabling applications such as predicting cancer adjustment, mental health outcomes after bereavement, and differentiating suicidal from non-suicidal patients. Complementing these clinician-facing applications, \citeasnoun{vanBuchem2022AIPREM} introduced the AI-PREM, an open-ended patient experience questionnaire paired with an NLP pipeline that combines topic modeling and sentiment analysis to thematically cluster short patient narratives into actionable summaries for clinicians. These examples underscore the role of linguistic feature analysis in diagnosis, management, and prognosis. However, Clinical NLP faces challenges in \textit{clustering} tasks, as noted by \citeasnoun{Sheikhalishahi2019} and \citeasnoun{ghosh2024natural_cluster}, including difficulties in processing clinical notes, ambiguities, context-dependent, and domain specific knowledge.

Recent advancements emphasize \textit{patient-narrative} data, aligning with this study’s focus. \citeasnoun{ohno2025japanese} developed a BERT-based model for monitoring symptoms from patient interviews in a Japanese hospital, achieving superior performance despite limitations like data scarcity and single-source training. Furthermore, \citeasnoun{alon2025decision} analyzed Hebrew clinician speech using word frequency and Large Language Models (LLMs) to uncover cognitive paradigms, finding reliance on heuristics and intuitive reasoning, though limited by small datasets and lack of multilingual frameworks. Similarly, \citeasnoun{Somani2023} used an AI pipeline combining BERTopic, dimensionality reduction, and spectral clustering to organize 10,233 patients' statin-related Reddit posts into six overarching themes, and then applied a pretrained BERT sentiment model, which found the discussions were predominantly neutral to negative. 
Overall, literature shows Clinical NLP’s potential to transform healthcare by extracting insights from both patient and clinician narratives, yet challenges persist due to privacy concerns, linguistic variability, and limited annotated datasets \cite{raj2023policy,han2024neural,bak2025ethical}. 
This study addresses these issues by exploring neural topic modeling for coherent, patient-centered topics, promoting reproducibility through shared and anonymized datasets and models, and investigating multilingual approaches using specialized embeddings and translation tools to advance linguistically inclusive Clinical NLP. 

\section{The Dataset - Patient \revision{Interviews}}

\label{sec:dataset}
The data we use consists of 13 anonymized, \revision{manually} transcribed \texttt{.docx} files (I0 to I12), each corresponding to a different \revision{pancreatic cancer} patient. As mentioned in the introduction section, the dataset is provided by the \revision{Erasmus MC, as the interviewees were patients of the Erasmus MC. More details about the data can be found in} \citeasnoun{griffioen2021bigger}.
%
\revision{
The interviews were conducted with 13 patients and 13 family members/supporters. The interviewing researcher was a service designer of TUDelft, doing her PhD on this DCS project.} 

The dataset contains interviews in which the patients discuss their experiences with the disease, such as how they were diagnosed, their emotional struggles, coping mechanisms during treatment, and other personal reflections. The length of each document varies, with the shortest document containing 5,596 words (approximately 34,000 characters), and the longest document containing 12,875 words (approximately 58,000 characters). The total collection size is 132,772 words.

\revision{The interviews were semi-structured, implying that the interviewer may elaborate on the answers given by the respondent. The interview agenda consists of 29 questions, ranging from ``how did the treatment go'' to ``what were your three main worries'', but due to the free nature of the conversations, the resulting interviews do not follow a fixed template. The only other available meta-data was a timestamp, e.g. ``maandag 7 oktober 2019, 02:21:36". This is not directly relevant to the topical analysis, so we leave it out from our data processing. } 
\revision{In our work, we use this dataset as provided, without the metadata and the interview agenda, and apply additional standard preprocessing steps, including stop words list and speaker label removal, conversion to plain text, and translation, etc. (more details in Section~\ref{sec:preprocessing}). No new annotations were introduced.
While the dataset is of high quality due to its curated origin, we acknowledge that its characteristics (e.g., size, cancer domain specificity) may limit generalizability. We discuss this further in Section~\ref{sec:discussion}.}

\begin{figure}[t]
  \centering
  \begin{minipage}{0.48\textwidth}
    \centering
    \includegraphics[width=\linewidth]{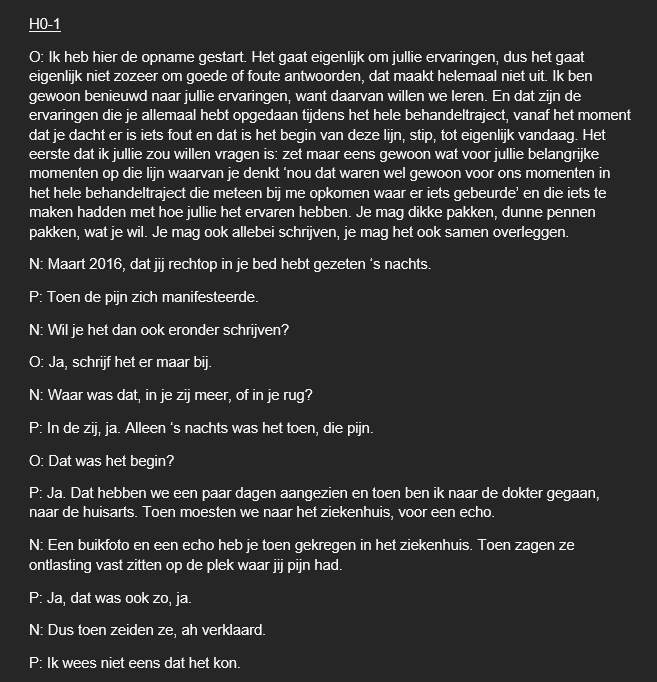}
    \label{fig:image1}
  \end{minipage}
  \hspace{0.02\textwidth}
  \begin{minipage}{0.48\textwidth}
    \centering
    \includegraphics[width=\linewidth]{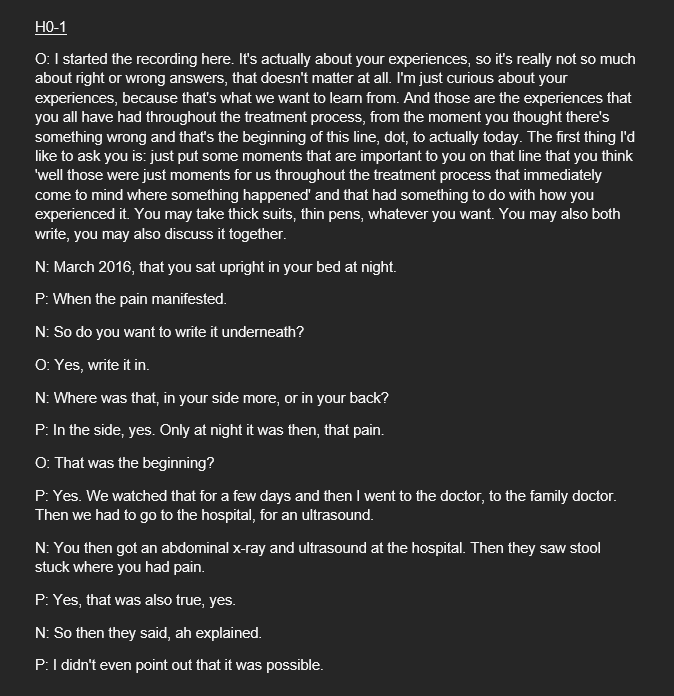}
    \label{fig:image2}
  \end{minipage}
  \caption{Snippets of the I0 interview in Dutch (left image) and English translation (right image). \revision{The interviewer refers to a  paper with the drawing of the care path up to that point. This was made on draft paper and later discarded.}}
  \label{fig:sample-text-interview}
\end{figure}

Each document features three different speakers: the patient, the researcher who conducts the interview, and the ``naaste", a Dutch word which may be roughly translated in English to \revision{``significant other" (the term used in the original paper)} or ``close relation", whose role during the interviews is to offer an outside perspective on the patient's cancer journey and emotional support. Each line within the document is marked with a capital letter which represents who is speaking: P for the patient, N for the ``naaste", and O for the interviewer. An example is shown in Figure~\ref{fig:sample-text-interview}. \revision{We used all data together: the patient, relative, and researcher utterances. The three speakers share topics throughout the conversation and are therefore modelled together.}

Moreover, all texts preserve marks of orality, such as hesitations, repetitions, and informal speech. While this format reflects the emotional nature of the interviews and provides a better insight into the patient's experiences, it also introduces potential complications in the \textit{preprocessing} and analysis stages, which are thoroughly discussed and analyzed in the upcoming dedicated section.

Another challenge lies in the language of the documents: all interviews are fully conducted in Dutch. This presents a potential barrier, as 
many \revision{pretrained} domain-specific embedding models tend to perform best on English-language data or are even exclusively English (especially in the biomedical domain), which may impact the quality and consistency of the results when working with Dutch transcripts. 

We examine structural and linguistic challenges in detail in the upcoming dedicated sections, which explore workarounds and compromises for not only preserving the main essence and important aspects of the original transcripts but also for producing accurate and easily interpretable results by showcasing various experiments and alternative options.

\begin{figure}[t!]
    \centering
    \includegraphics[width=1\textwidth]{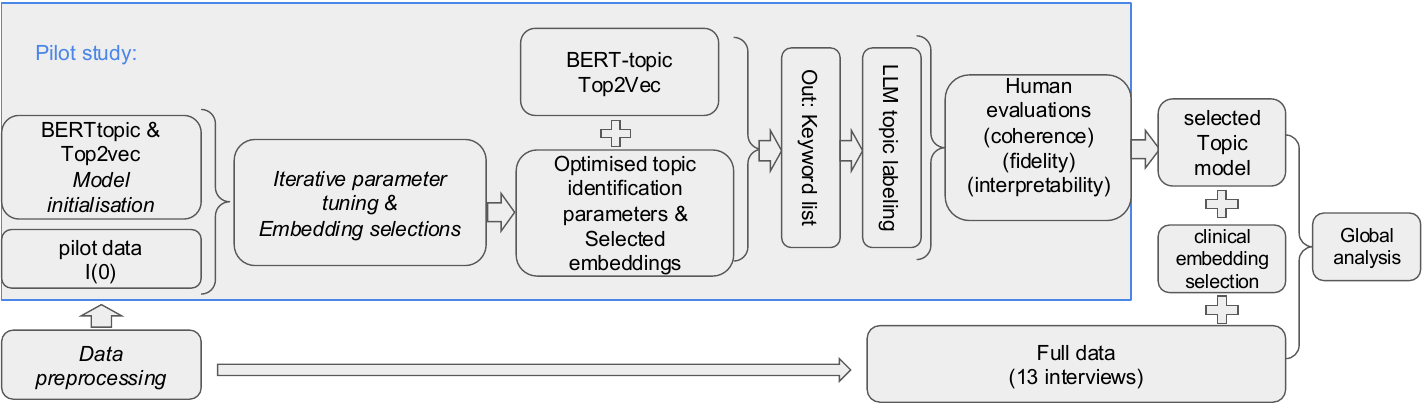}
    \caption{Framework Design of Overall Experimental Investigation}
    \label{fig:pipeline-fig}
\end{figure} 

\section{Design of Methods}
The goal of this study is to identify key topics and themes present in patient \revision{interview} data and examine their relevance within the context of the interview, in order to determine whether state-of-the-art neural topic modeling techniques can be useful for providing patient feedback to medical staff. 
To achieve this, BERTopic and Top2Vec are applied on a single anonymized cancer patient interview at a time, with the pilot-study interview being \texttt{Interview I0}, and their outputs are evaluated in order to determine their coherence and relevancy in a medical context, as shown in the system development framework in Figure \ref{fig:pipeline-fig}.

Out of available topic modeling methods, we first select \textbf{BERTopic} because of its flexibility in choosing an embedding model, as well as its capability to capture context and synonyms, which is important for medical jargon and marks of orality.\footnote{available at \url{https://github.com/MaartenGr/BERTopic}} For example, the word ``patient" has a different meaning when used as a noun, referring to a person receiving medical care, compared to its use as an adjective, where it describes the quality of being tolerant. This distinction is especially important in medical interviews, where context determines whether terms describe people, conditions, or behaviors. Moreover, BERTopic does not require manually setting the number of topics in advance, which is useful if the number of themes that may emerge from a corpus is unknown. This is achieved through its use of HDBSCAN as clustering algorithm, which determines the number of clusters based on the distribution and density of the embedded documents. 
For comparison purpose, we select \textbf{Top2vec} as an alternative neural topic model.\footnote{available at \url{https://github.com/ddangelov/Top2Vec}} 
While BERTopic is designed to use transformer-based models to better capture contextual nuance, Top2Vec originally used Doc2Vec embeddings exclusively, and was later extended to support more powerful models like Universal Sentence Encoder \cite{universal}. 

As shown in Figure \ref{fig:pipeline-fig}, from bottom-left to right, the dataset is firstly preprocessed and chunked into segments (the exact number varies per experiment and file) in order to preserve narrative coherence while also ensuring compatibility with the embedding models in terms of context length. 
In the Pilot-study, both BERTopic and Top2Vec are tested on the default settings in order to check for preliminary issues and to further experiment with the parameters of the models based on the initial outputs. 
Moreover, we tested several different embedding models as backbones at the ``iterative parameter tuning and embedding selections'' phase, but due to the differences between BERTopic and Top2Vec, the focus is placed on a common embedding model, namely \textbf{all-mpnet-base-v2} \cite{reimers2020allmpnetbasev2}, in order to fairly evaluate their performance on as much common ground as possible. 
During this phase, the parameters of each model, as well as the parameters that control the clustering and dimensionality reduction processes, are experimented with and modified across several runs to determine which settings yield the best results.

Subsequently, the optimized topic identification parameters and selected embeddings (available within the two neural models) are used for both neural topic models to produce keyword lists. We use an LLM to summarize topics from the keyword list into a descriptive label. Then human evaluations are carried out on topic model selection by looking into aspects of \revision{coherence and usefulness, representativeness to the interview content, and helpfulness of the keywords}.

For analyzing the entire interview data, we will further explore clinical domain embedding models, beyond the default embeddings from BERT-topic and Top2Vec themselves, together with the selected topic model framework from the pilot study, for a global analysis, in Section \ref{sec:global}.


\section{Pilot-study and Optimizing Topic Models}

\subsection{Data Preprocessing} \label{sec:preprocessing}

As stated in Section \ref{sec:dataset}, the source data consists of 13 anonymized cancer patient interviews stored as \texttt{docx} files, which contain speaker labels (P for patient, N for \revision{the significant other}, and O for the interviewer), structural markings such as headers and internal identifiers, as well as marks of orality, such as \revision{disfluencies} and repetitions. 
Our preprocessing pipeline is listed in Table~\ref{tab:preprocessing} with examples, including conversation to plain text conversion, document translation, speaker label removal, section header removal, contraction expansion, and removing custom stop words list.

For model development and pilot-testing, we first apply each model to a single preprocessed interview (I0) and test various configurations to explore how different parameter choices and data representations influence the quality and interoperability of the resulting topics, detailed in next sections separately for two neural models.

\begin{table}[t]
\centering
\caption{Preprocessing Pipeline}
\label{tab:preprocessing}
\renewcommand{\arraystretch}{1.3} 
\begin{tabular}{|p{4cm}|p{9.5cm}|}
\hline
\textbf{Preprocessing Step} & \textbf{Purpose} \\
\hline
\revision{Conversion} to plain text & Converting the document to a clean \texttt{.txt} file by using the \texttt{python-docx} package \cite{python-docx}\\
\hline
Document Translation & Translation from Dutch to English with DeepL \cite{deepl} for the application of domain-specific backbone models and interpretation by non-Dutch speakers. \\
\hline
Speaker Label Removal & Remove speaker tags (e.g., \texttt{P:}, \texttt{N:}) \revision{These tags are used as prefixes to the utterance to indicate the speaker. These are not meaningful tokens for the topic models, which are embeddings-based and use word meaning as central building block. These prefixes also do not contribute to the topical analysis of the interviews, but have a purely structural meaning.}
\\
\hline
Section Header Removal & Remove structural markers (e.g., \texttt{I0-1}) that do not carry any semantic meaning. \\
\hline
Contraction Expansion & Prevent broken tokens (e.g., \textit{wasn}, \textit{t}) by expanding contractions (e.g., \textit{wasn't} to \textit{was not}). \\
\hline
Custom Stop Words List & Remove uninformative and meaningless words (e.g., ``uh'', ``yeah'', "says"). \\
\hline
\end{tabular}
\end{table}

\subsection{Model Development and Optimization: BERTopic}
\label{ref:bertexp}

To manage the length of \revision{interview} transcripts and ensure compatibility with embedding model input limits, we apply a sentence-based chunking strategy. After processing, the text is split into sentences using a regex-based approach, chosen for its simplicity and sufficient accuracy given the well-punctuated nature of the transcripts. These sentences are then grouped into fixed-size chunks, and different chunk sizes are tested to assess their impact on topic generation and coherence. 
Table \ref{tab:chunkres} shows that the number of chunks directly influences the number of topics produced. 

\begin{table}[t]
\centering
\caption{Effect of Sentence Chunk Size on Number of Chunks and Topics for Interview I0}
\label{tab:chunkres}
\begin{tabular}{|l|l|l|}
\hline
\textbf{Sentences Per Chunk} & \textbf{Chunks} & \textbf{Topics} \\ \hline
5                            & 172             & 17              \\ \hline
6                            & 144             & 16              \\ \hline
7                            & 123             & 12              \\ \hline
8                            & 108             & 9               \\ \hline
\end{tabular}
\end{table}

We start with BERTopic's default settings and the all-mpnet-base-v2 embedding model, briefly testing alternatives like MiniLM-L6-v2, which we discard due to token limits and inconsistent outputs. We adjust UMAP parameters (\texttt{n\_neighbors}, \texttt{min\_dist}, \texttt{n\_components}) to balance topic separation and cohesion, and tune HDBSCAN settings such as \texttt{min\_cluster\_size} and \texttt{min\_samples} to control granularity and noise. The vectorizer is optimized to include bigrams, improving interpretability by capturing multi-word expressions common in clinical narratives. Finally, we experiment with \texttt{min\_topic\_size}, finding the default value (10) best preserves both broad and specific themes without excessive fragmentation.

\subsection{Model Development and Optimization: Top2Vec}
\label{ref:top2vecexp}

Similar to the tuning process employed for BERTopic, Top2Vec is optimized through iterative refinements of key parameters. This time, however, we start from the parameters of the already-tuned BERTopic model in order to see how Top2Vec behaves differently under close-to-identical conditions to BERTopic. Most of the tuning process that follows involves altering the parameters passed to the UMAP and HDBSCAN components that are internally used during the dimensionality reduction and clustering. We evaluate the same UMAP parameters explored during BERTopic tuning, namely \texttt{n\_neighbors} \texttt{min\_dist} and \texttt{metric}, to observe their influence on topic formation. Similarly, we modify the HDBSCAN parameters such as \texttt{min\_cluster\_size} and \texttt{min\_samples} in order to balance specificity and generalization within the topic clusters.

In our experiments, the default Top2Vec settings cause clustering errors, but tuning UMAP and HDBSCAN parameters resolves this and produces meaningful topics. Unlike BERTopic, Top2Vec \textit{does not} allow custom n-gram or stop-word settings, \revision{limiting keyword flexibility. 
After} iterative refinements, the optimized Top2Vec models generate topics suitable for comparative analysis with BERTopic.

\subsection{Topic Labeling with an LLM on Keyword Lists}
\label{sec:label}

Because the output topics from both models are essentially lists of keywords, they do not hold an interpretable meaning at first glance. To make sense of them, each topic must be labeled according to its semantic coherence. Traditionally, this involves manually looking at the top keywords for each topic, along with the sample documents that determined the formation of the topic. Labels are usually chosen based on the literal meaning of the keywords, along with the context found in the actual text, in order to make them interpretable and meaningful for human readers. The goal is for the label to be as meaningful as possible, especially in a healthcare setting where clarity and relevance are essential.

In order to achieve this, we use a large language model (LLM) to label the topics in order to automate the process. The specific LLM model that we use for this task is OpenAI's \textbf{GPT-4o mini} model, because it is a powerful yet cost-efficient and lightweight model that shows strong performance across a range of evaluation metrics \cite{openai2024gpt4omini}. At first, we only pass the topic keywords to the model to see if the resulting topic labels would be cohesive enough without the need for representative documents.

The model generates descriptive labels only based on the topic's keywords, which produces mixed results. While some topic labels are good representations of the actual topics, others are unreliable because they lack the context behind the actual keywords. For example, one of the output topics produced by BERTopic, with some of the top keywords being ``size 19, tricky, 25" received the label ``Size Discussion in Medical Context", which has nothing to do with the actual context behind the topic. The absence of document context substantially impacts the quality of the generated labels, which is an impediment for individual-interview analysis. We therefore adapt the prompt \textit{including document snippets} as additional \textit{contextual} evidence to help interpret the topics more accurately. The final prompt is the following:

\begin{quote}
    ``You are an AI that labels discussion topics, from a cancer interview, for a software that allows doctors to browse through medical files without the need to read them from start to finish. Given the following keywords and sample documents, provide a clear and specific topic label, focusing mainly on the keyword list and using the document snippets as supporting context rather than a baseline. Only type the topic label and nothing else:".
\end{quote}


\section{Pilot-study Results and analysis}

\subsection{BERTopic Results}
\label{sec:bertresults} 

We first evaluate BERTopic on the preprocessed content of interview I0. The final model uses the \texttt{all-mpnet-base-v2} embedding model for semantic encoding, UMAP for dimensionality reduction, and HDBSCAN for clustering. 
Using the optimized configuration, BERTopic generates a total of \textbf{17 topics}, each representing a relatively coherent cluster of semantically related segments within the interview. These topics span a wide range of themes, from procedural experiences and emotional reflections to logistical concerns and treatment decision-making, deeming this output suitable for a clinically-oriented software that clinical staff can use to quickly analyze patient data. 
A comprehensive overview of the topics, along with their automatically generated labels and keywords, is presented in Table \ref{tab:bertopic_full_labels}. The \texttt{all-mpnet-base-v2} embedding model manages to produce rich semantic representations that help to distinguish nuanced narratives.

The model output results in topics that are not only clinically relevant but also reflective of the patient's emotional and experiential journey through cancer treatment. For instance, \textbf{Topic 16} describes the patient's experiences during their FOLFIRINOX \cite{nci_folfirinox} chemotherapy treatment, highlighting concerns such as neuropathy and treatment planning. Insights like these can serve as valuable feedback for clinical staff, offering a patient-centered perspective on how individuals are coping with the physical and psychological effects of specific therapies. On the other side of the spectrum, \textbf{Topic 8} offers the patient's logistical challenges and communication-related experiences, particularly in coordinating appointments and interactions with medical staff at Erasmus Hospital in Rotterdam. These types of topics can offer useful feedback to the hospital itself for improving internal processes, such as appointment coordination, patient communication, and overall administrative support. By surfacing these issues from patient narratives, the model provides actionable insights that can contribute to a more patient-oriented care experience.  

The pilot study also shows limitations of BERTtopic. 1) Some topic keywords contain duplicate words due to the \textbf{ngrams} parameter. For example,in the keywords list of \textbf{Topic 7}, the word ``port cath" appears as three different entries: ``port", ``cath", and ``port cath". While the inclusion of bigrams increases contextual richness, it can also lead to partial redundancy when both a phrase and its constituent words appear independently in the keyword list. Although this does not substantially impact interpretability, it can affect visual clarity and compactness of the topic's summary. On the other hand, this overlap may be beneficial when dealing with noisy or varied language, as it may help ensure that key terms are captured even when they appear in different forms across the corpus. This error could potentially be fixed in a post-processing process, or even within the vectorizer settings with extra tuning. Future iterations could involve filtering out redundant unigrams when a high-confidence bigram is present, though this would need to be balanced against the risk of losing relevant variations in phrasing. 2) Another issue is overlapping themes spanning different topics. 
For example, port-a-cath procedures are discussed in two different topics: \textbf{Topic 4} and \textbf{Topic 7}. 
While the lists of keywords are different, meaning that while port-a-cath is discussed, this could reflect a true conversational shift around the same object of interest, they still share a common theme, which could be a sign of over-segmentation driven by overly sensitive clustering parameters. 
Although this could become an issue, manual document analysis proves that the two topics are distinct enough to be classified in individual clusters. Therefore, this occurrence is not a substantial issue, especially for this preliminary comparison phase, but rather proof that the model is capable of identifying nuanced themes, even from within the same broader topic.


\begin{table}[t]
\centering
\caption{ Output of the Optimized BERTopic Model (Interview I0): pilot study}
\resizebox{\textwidth}{!}{
\scriptsize
\begin{tabular}{|l|p{5cm}|p{8cm}|}
\hline
\textbf{Topic ID} & \textbf{Topic Label}                                                                                            & \textbf{Top 15 Keywords}                                                                                                                                        \\ \hline
0                 & Experiences and Challenges Navigating Patient Passes at Daniel den Hoed Cancer Center                           & weird, white, pass, understand, den, hoed, daniel, daniel den, den hoed, patient, outside, room room, team, notice, waiting                                  \\ \hline
1                 & Experiencing Fear and Anxiety During Medical Examinations Involving Tubes and MRI Scans                         & period, easy, lying, scary, tubes, throat, mri, tumor marker, marker, examination, prepared, tumor, rest, sorry, ultrasound                                  \\ \hline
2                 & Radiation Treatment Timeline: Delays, Scheduling, and Scans from June to November                               & radiotherapist, radiation treatments, june, months, months scan, end november, guus, guus meeuwis, meeuwis, radiation, december, 21, follow, november, radio \\ \hline
3                 & Timeline of Malignant Diagnosis and Hospital Visits Including MRI and Ultrasound Examinations                   & malignant, april, write, timeline, place hospital, pretty, mri, walking, town, ultrasound, start, work, recording, raise, hospital hospital                  \\ \hline
4                 & Challenges and Experiences with Blood Draws and Port-a-Cath Access in Cancer Treatment                          & prick, cath, port, port cath, blood, needle, markers placed, hand, day, placed, markers, poked, difficult, puncture, puncture room                           \\ \hline
5                 & Discussion on Communication and Decision-Making in Pancreatic Cancer Treatment Conversations                    & eating, talk, cancer, list, clear, pancreatic cancer, pancreatic, certainly, eventually, conversation, surgeon, helps, time exciting, money, evening         \\ \hline
6                 & Encouraging Patients to Ask Questions and Address Concerns During Appointments                                  & especially, calls, questions, concerns, monday, head, difficult, mean, appointments, ask, involved, asked things, doctor going, reach, regular               \\ \hline
7                 & Challenges and gaps in understanding port-a-cath placement and related surgical procedures.                     & brand, surgery, kind information, size, port, port cath, cath, example, information, 21, rotterdam, work, happens, eligible, drive                           \\ \hline
8                 & Patient's experience coordinating appointments and communication with doctors at Erasmus Hospital in Rotterdam. & forget, wednesday, doctor doctor, hair, date, friday, send, rotterdam, surgeon, erasmus, parking, specialized, hospital came, hospital want, went hospital   \\ \hline
9                 & Discussion of tumor markers and the timeline of metastases detection and monitoring.                            & metastases, months, heard, tumor marker, marker, lot, year, tumor, spots, normal, lot googled, information, months blood, sampling, blood sampling           \\ \hline
10                & Impact of 2017 Cancer Cure on Patient's Life and Recovery Experience                                            & 2017, cure, year, hands, true course, opinion, gee, intense, 11, husband, home, took long, clear, certainly, rest                                            \\ \hline
11                & Bowel Test Results and Persistent Pain Leading to Further Medical Referral                                      & taken, test, feeling, bowel test, piece, bowel, poking, worked, referred, pain, monday, showed, away, read, touch                                            \\ \hline
12                & Family Doctor Interactions and Patient Concerns During Cancer Treatment Interviews                              & family doctor, family, interview, grumpy, doctor family, worry, guys, time time, time doctor, long hair, hair, surgeon, nurse, definitely, puncture          \\ \hline
13                & Awkward Appointment Experiences and Scan Discussions Over Two and a Half Years                                  & appointments, sigh, showing, half years, appointments new, years, look, scans, came scan, awkward, learned, exactly question, experts, cd, forward           \\ \hline
14                & Discussion of treatment options and choices made with the doctor during cancer care.                            & ask doctor, woolly, place place, ask, goodbye, familiar, doctor actually, choice, discussed, options, effective, hospital ask, time heard, kept, want want   \\ \hline
15                & Discussion on managing interruptions and planning during cancer treatment conversations.                        & prepared, stop, coming, clearly, time speak, stop moment, plan, mention, fine, going, feel, moment, right, people, happy                                     \\ \hline
16                & Concerns about neuropathy during FOLFIRINOX treatment process and course options discussed.                     & neuropathy, concerns, folfirinox, told, treat, courses, left, process, treatment process, folfiri, disease, huge, want want, write, treatment                \\ \hline
\end{tabular}
}
\label{tab:bertopic_full_labels}
\end{table}

\subsection{Top2Vec Results}
\label{sec:topvecresults} 

The final configuration of the model, including the UMAP and the HDBSCAN parameters, largely mirrors those used for BERTopic. The only major difference is the use of \texttt{min\_count} parameter in place of \texttt{min\_df}, which serves a similar function by controlling the minimum frequency a word must appear in the corpus to be considered for the clustering process. The embedding model used by default in Top2Vec is \texttt{all-MiniLM-L6-v2}, as confirmed by the verbose output during the fitting process. 



Using the optimal configuration, Top2Vec generates a total of \texttt{18 topics}, which is two topics more than the output generated by BERTopic. These topics are generally cohesive and interpretable, and could be reliably mapped back to the original chunks used during the model training. The final model output, along with labels and keyword lists, is presented in Table \ref{tab:top2vec_full_labels}. In several cases, the model produces topic clusters that closely resemble those found in the BERTopic output. For instance, two distinct topics relating to port-a-cath procedures emerge, similar to those identified by BERTopic, with keywords reflecting different aspects of the same clinical theme. However, not all topic clusters capture the broader context with the same accuracy. In one case, a conversation surrounding needle sizes is isolated into a broader, more general topic, namely \textbf{Topic 18}, but the model fails to recognize that the discussion is part of a larger conversation about needle types and procedures, as the interview transcripts suggest. As a result, the topic is reduced to a generic theme about ‘‘experiences and information", with keywords focusing on different medical and non-medical terminology and procedural words, such as ``asked" and ``conversation", rather than a cohesive and concrete theme.
In general, the model produces interpretable results that align reasonably well with human-labeled themes. The ability to trace topic clusters back to specific document chunks makes manual evaluation possible and useful. 

\begin{table}[t]
\centering
\caption{Output of the Optimized Top2Vec Model (Interview I0): pilot study}
\resizebox{\textwidth}{!}{
\scriptsize
\begin{tabular}{|l|p{5cm}|p{8cm}|}
\hline
\textbf{Topic ID} & \textbf{Topic Label} & \textbf{Top 15 Keywords} \\ \hline
0  & Patient Experiences and Emotions During Cancer Appointments and Treatment Environments & suddenly, conversation, room, happened, exciting, talk, appointment, story, hadn, later, move, cry, experiences, felt, was \\ \hline
1  & Challenges and Experiences During Cancer Diagnosis and Treatment Journey & mri, surgeon, surgery, appointment, patient, doctor, ultrasound, tumor, hospital, appointments, metastases, scan, cancer, examination, malignant \\ \hline
2  & Patient Experiences with Cancer Treatments, Cures, and Recovery Challenges & cure, cures, treatments, patient, treatment, pain, surgery, doctor, metastases, rest, puncture, appointment, no, yes, mri \\ \hline
3  & Patient Experiences and Challenges During Cancer Treatment, Appointments, and Surgical Procedures & treatment, treatments, patient, appointment, doctor, surgery, chemo, cure, cures, experiences, metastases, cancer, appointments, concerns, experienced \\ \hline
4  & Interactions with Healthcare Professionals During Cancer Diagnosis and Treatment & doctor, surgeon, hospital, patient, appointment, nurse, nurses, surgery, appointments, interview, ultrasound, examination, maybe, secretary, coach \\ \hline
5  & Patient Experiences with Doctor Communication and Appointment Management in Cancer Care & appointment, patient, hospital, doctor, appointments, nurse, call, conversation, calls, nurses, surgery, suddenly, treatment, talk, surgeon \\ \hline
6  & Timeline of Cancer Diagnosis and Treatment: Metastases, Tumor Markers, and Patient Experience & metastases, tumor, cancer, malignant, chemo, mri, patient, appointment, surgery, treatments, months, march, timeline, treatment, weeks \\ \hline
7  & Patient Experiences with Doctors and Hospitals During Cancer Diagnosis and Treatment & doctor, hospital, appointment, patient, surgeon, appointments, nurse, nurses, surgery, ultrasound, mri, tumor, malignant, metastases, examination \\ \hline
8  & Patient Experiences and Communication with Medical Professionals at Erasmus Hospital for Cancer Treatment & erasmus, appointment, doctor, surgeon, surgery, hospital, patient, appointments, treatment, examination, tumor, nurse, mri, puncture, treatments \\ \hline
9  & Patient Experience with Radiation Treatment and Follow-Up Appointments for Cancer Management & radiation, radiotherapist, radio, patient, treatment, treatments, appointment, surgery, cancer, metastases, tumor, doctor, cure, mri, surgeon \\ \hline
10 & Challenges and Experiences with Blood Draws and Port-a-Cath Procedures in Cancer Treatment & needle, puncture, poked, patient, blood, ultrasound, mri, prick, tubes, appointment, scan, port, treatments, surgery, examination \\ \hline
11 & Exploring the Role of Turmeric and Patient Questions in Cancer Treatment Decisions & turmeric, cures, treatments, cure, certainly, only, no, concerns, definitely, doctor, maybe, treatment, always, probably, surgeon \\ \hline
12 & Navigating Hope and Treatments in Pancreatic Cancer: A Patient's Journey and Conversations & cancer, pancreatic, tumor, chemo, metastases, malignant, patient, cure, treatments, cures, treatment, doctor, hope, surgery, bowel \\ \hline
13 & Experiences and Concerns Surrounding Port-a-Cath Placement and Blood Draw Procedures & surgery, port, puncture, patient, needle, cath, surgeon, tubes, nurse, hospital, obviously, fact, appointment, nurses, operation \\ \hline
14 & Bowel and Pancreatic Cancer Diagnosis Journey: Pain, Tests, and Treatment Experiences & bowel, pancreatic, patient, puncture, hospital, ultrasound, doctor, pain, appointment, mri, surgery, examination, scan, tumor, surgeon \\ \hline
15 & Patient Experiences with Chemotherapy Options and Neuropathy Management in Cancer Treatment & treatments, neuropathy, treatment, cure, doctor, patient, surgery, cures, surgeon, folfirinox, mri, chemo, tumor, cancer, pain \\ \hline
16 & Navigating Appointments and Treatment for Pancreatic Cancer at Rotterdam Hospital & rotterdam, hospital, appointment, erasmus, patient, appointments, surgery, doctor, pancreatic, surgeon, tumor, interview, treatment, experiences, examination \\ \hline
17 & Patient Experiences and Concerns Regarding MRI and Scan Examinations in Cancer Care & scan, mri, ultrasound, certainly, examination, appointment, yes, patient, radio, no, ask, probably, radiotherapist, well, asked \\ \hline
18 & Patient Experiences and Information Gaps During Surgical Appointments and Examinations in Oncology & size, patient, surgery, questions, doctor, information, experiences, examination, nurse, nurses, surgeon, experienced, appointment, conversation, asked \\ \hline
\end{tabular}
}
\label{tab:top2vec_full_labels}
\end{table}

\subsection{Model Evaluation and Comparison with Human Perspectives}
\label{sec:modelanalysis}
To evaluate the coherence, relevance, and clinical interpretability of the topics produced by BERTopic and Top2Vec, we conducted a small-scale user study involving three volunteer participants. Each participant was provided with the anonymized cancer interview I0 and asked to read it carefully. They were then asked to answer five questions for each model output. The full survey can be found in Appendix \ref{ref:appendixa}. While the volunteers are not clinical experts, the task of judging the coherence and contextual relevance of the topics does not require specialized expertise and can be meaningfully performed by general readers with enough context from the source material. \revision{We focus our analysis on the responses to Q1 and Q3, because Q2 required reviewing of the full interview, which was more challenging and therefore less conclusive. In response to Q1, p}articipants rated each topic on a scale from 1 to 5 based on its coherence and usefulness (Q1), and \revision{in response to Q3, they} evaluated the associated top 15 keywords \revision{for helpfulness to understand what the topic was about}. They were also asked to identify any essential themes that the models may have missed (Q4). The outputs that they were provided with were Tables \ref{tab:bertopic_full_labels} and \ref{tab:top2vec_full_labels}. They did not have access to the representative documents from each topic, but they were allowed to go back to the interview in order to check whether the topic label was, in fact, describing a topic that was talked about during the interview. 

\begin{figure}[t]
    \centering
    \includegraphics[width=0.7\linewidth]{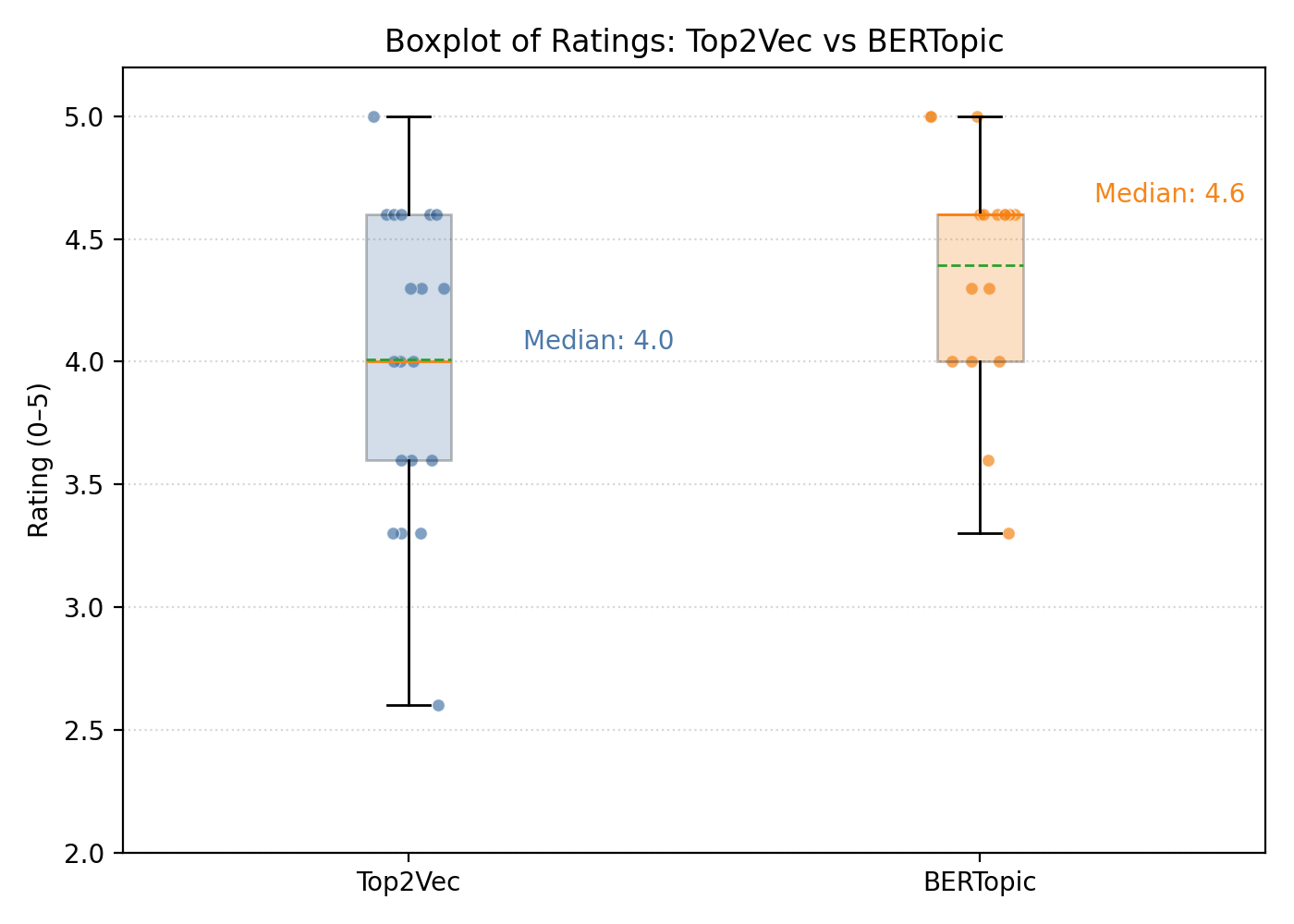}
    \caption{Distribution of ratings for Top2Vec and BERTopic topics \revision{in response to  Q1: coherence/usefulness of the topics}, averaged over 3 raters. Note that Top2Vec has generated 19 topics and BERTopic has generated 17 topics.}
    \label{fig:boxplot_Q1}
\end{figure}

\begin{figure}[t]
    \centering
    \includegraphics[width=0.7\linewidth]{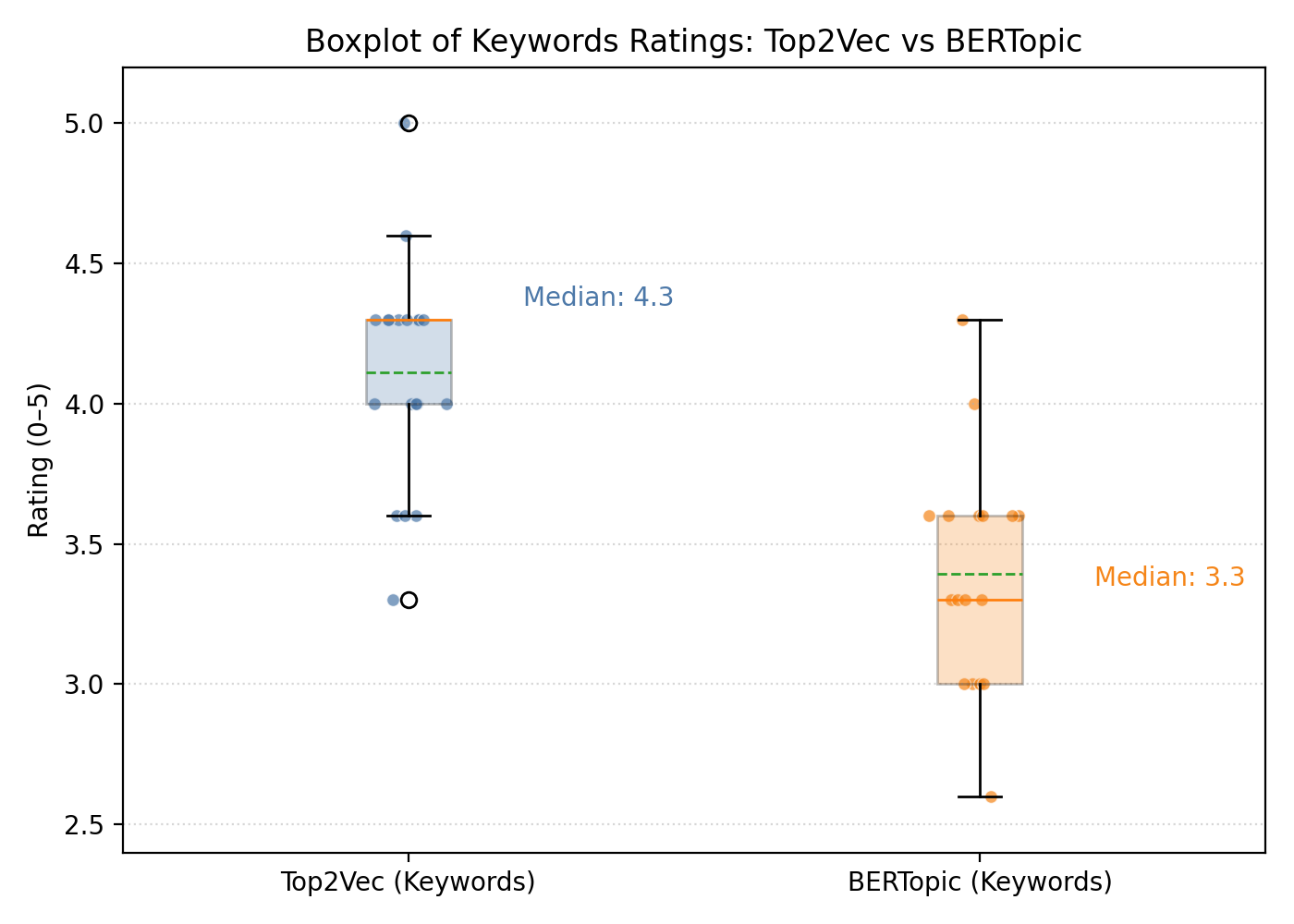}
    \caption{Distribution of ratings for Top2Vec and BERTopic keywords \revision{in response to Q3: helpfulness of the terms}, averaged over 3 raters. Note that Top2Vec has generated 19 topics and BERTopic has generated 17 topics.}
    \label{fig:boxplot_Q3}
\end{figure}

The results for Q1 (topic coherence and usefulness) are in Figure~\ref{fig:boxplot_Q1}. It shows that the ratings for BERTopic are on average higher and there are fewer topics that score very low compared to Top2vec. Overall, BERTopic performed strongly, with 12 out of the 17 topics receiving a coherence score of 4.0 or higher. The results for Q3 (keyword descriptiveness) are in Figure~\ref{fig:boxplot_Q3}. \revision{While Top2Vec achieved lower \textit{topic coherence/usefulness} scores on average than BERTopic, Top2Vec performed better in terms of \textit{keyword helpfulness} ratings:} 15 out of 19 topics received a keyword score of 4 or above, indicating that participants generally found the keywords descriptive and relevant to the topic content. A common point of feedback from the participants\revision{, however,} was that some topics appeared to overlap in content, such as between \textbf{Topics 4, 5, and 7}. \revision{Apparently, the 
fact that BERTopic scores higher on coherence/usefulness (Q1, Figure~\ref{fig:boxplot_Q1}), indicates that the set of terms does not have to be perfect (Q3, Figure~\ref{fig:boxplot_Q3}) without harming the topic model evaluation.} 
This shows that while Top2Vec extracted more \textit{potential} topics than BERTopic, it might fail to capture nuanced themes, which are crucial for the primary goal of this research.

For Question 2, which asks how well the extracted topics represent the content of the interview overall, the BERTopic model was ranked higher than the Top2Vec model, with the added mention from all three of the evaluation participants that, while both models manage to extract mostly cohesive topics from the I0 interview, the BERTopic model's output is way more \textit{precise} and contained \textit{less overlap}.

Reviewing the questions we designed:
\begin{itemize}
    \item Q1/Q2 is on the topic label (from LLMs) to the interview - Bert-topic wins, meaning Bert-topic is better at reflecting the interview topic.
    \item Q3 is on how well the extracted key word list (from TModel) supports the labeled topics (from LLM) - Top2Vec wins, meaning Top2Vec extracted keyword lists better connect with the summarized topic label.
\end{itemize}
From the overlapping (phenomenon) and cognitive (human reasoning effort) perspectives, we hypothesis that the more over-lapped keyword/theme list from Top2Vec can help annotators understand/infer the labeled topic better with less inference effort, because more isolated topic keywords (e.g. from BERT-topic) would need annotators to carry out more reasoning themselves to connect to the labeled topics.
This is comparison-wise. 
On the other hand, however, this does not necessarily mean that the extracted keyword list by Top2Vec are better reflecting the interview document in comparison to BERT-topic.

Taking the evaluation process, as well as the experimentation and results, into consideration, we decide to focus the analysis on BERTopic further. Not only did BERTopic demonstrate that it can consistently produce topics of high overall quality, but it also offers greater flexibility and extensibility because it is compatible with a wide range of publicly available embedding models. This opens up the possibility of integrating \textit{\textbf{clinically-oriented} embedding} models that may further enhance the model's ability to extract relevant, interpretable, and context-sensitive information from patient narratives.

\begin{figure}[t!]
    \centering
    \includegraphics[width=1\textwidth]{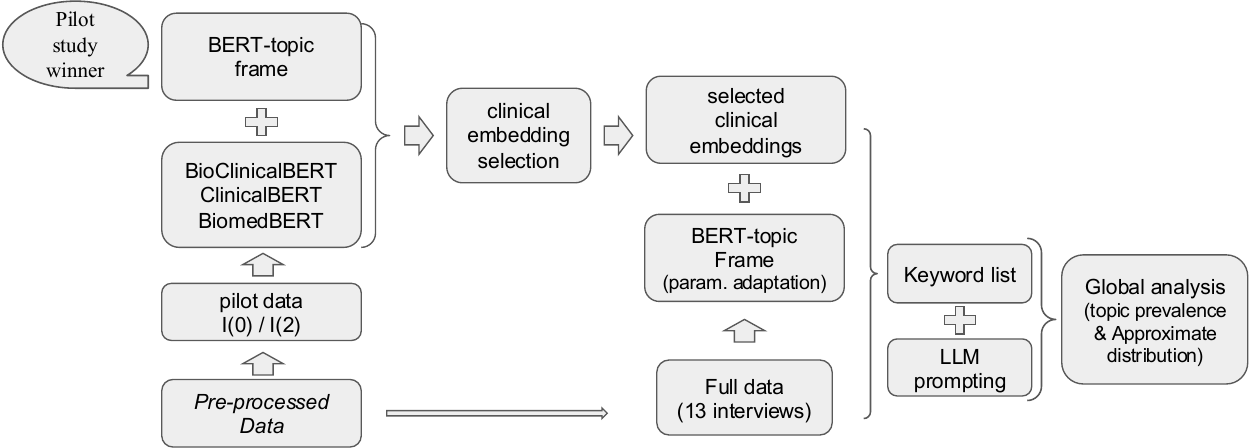}
    \caption{BERT-topic Deployment on all Interviews with Clinical LM Embeddings}
    \label{fig:bert-go-clinical-fig}
\end{figure}

\section{Investigating the Generalizability with Global Analysis}
\label{sec:global}

\subsection{Domain-specific Clinical/Medical Embedding Model Selection}
\label{sec:furtherexp}

To explore whether clinically oriented embeddings improve BERTopic's performance on patient \revision{interview} data, we experimented with three widely used models on I(0) interview: \textbf{BioClinicalBERT} \cite{bioclinical}, \textbf{ClinicalBERT} \cite{liu2025generalist-clinicalBert}, and \textbf{MSR BiomedBERT} (formerly PubMedBERT) \cite{pubmedbert} for domain-specific embedding model selection, as shown in Figure \ref{fig:bert-go-clinical-fig}. 
All experiments used the same BERTopic configuration as in Section \ref{ref:bertexp} to ensure consistency. \revision{The final outputs of all three embedding models are shown in Appendix \ref{ref:appendixb}. We qualitatively analyzed the output of the three models:}
\begin{itemize}
\item \textbf{ClinicalBERT} produced 17 topics initially, \revision{but we found some of these topics to be semantically incoherent. For example, conversations about needle sizes are grouped into a general-sounding topic regarding ``medical sizes". To reduce noise and improve topic quality, we increased the minimum document frequency (\texttt{min\_df}) to 3 and the chunk size from 6 to 7 sentences. This adjustment results in 15 topics, but these proved quite general or broad. To refine the topic boundaries, we reduced the dimensionality reduction setting \texttt{n\_components} from 10 to 8, which reduces the dimensionality of the UMAP projection and can help enforce tighter semantic clustering. While this maintains the 15-topic output and results in slightly more coherent themes, the quality of the extracted keywords suffers as a consequence. Redundancies and repeated terms appeared more frequently, making it harder to interpret the topics at a glance. Despite these issues, the output does seem to be an improvement over the initial BERTopic output found in Table \ref{tab:bertopic_full_labels}.}
\item \textbf{MSR BiomedBERT} also struggles with topic coherence to a greater extent than the previous models. For example, one of the topics, incorrectly labeled as ``Monitoring Eye Health and Treatment Progress in Cancer Care Discussions", includes representative document chunks that are not only short and uninformative, but one of the documents also includes the expression ``to keep an eye on", which is the reason for the misleading topic label. This means that the model struggles to create meaningful clusters, which leads to the LLM mislabeling the topic due to the lack of context and coherence within the representative documents and the list of keywords. This relatively weaker performance may be attributed to the nature of the dataset used to pretrain MSR BiomedBERT, which is probably less aligned with narrative-style patient data. 
\item In contrast, \textbf{BioClinicalBERT} consistently generated more interpretable and clinically meaningful topics, likely due to its pretraining on real clinical notes. Its outputs captured both technical medical experiences and patient perspectives with greater nuance, outperforming the other models in terms of coherence and relevance. These findings suggest that embedding models trained on clinical narratives provide \textit{stronger semantic representations} for this type of data.
\end{itemize}
Further tests on shorter interviews (such as I(2), which is the shortest in the dataset at approximately 5596 words) revealed that chunk size also plays a critical role: Reducing the chunk size to 6 sentences increases the number of topics to I(2), with a noticeable improvement in nuance and precision. The resulting topics capture more specific clinical moments and emotionally substantial statements, enhancing the interpretability and relevance of the output. This indicates that a dynamic chunking strategy, combined with clinically oriented embeddings such as BioClinicalBERT, may be essential for optimizing topic modeling across datasets with varying lengths and complexities.

\subsection{Global Analysis on All Interview Data}
\label{sec:globalanalysis}

\subsubsection{Model Setup}
As shown in Figure \ref{fig:bert-go-clinical-fig} (middle-horizontal), to gain deeper insights into the full dataset of 13 interviews, we applied topic modeling to the entire corpus rather than individual interviews, aiming to uncover overarching themes and recurring patterns. 
For this global analysis, we used \textbf{BioClinicalBERT}, identified in Section \ref{sec:furtherexp} as the most effective domain embedding model for producing coherent and clinically meaningful topics. 
Its domain-specific training on biomedical and clinical text makes it well-suited for capturing general themes. 
Unlike the per-interview approach, which prioritized fine-grained topics, this phase focused on \textbf{broader themes}. 

For model configuration, we increased the chunk size from 6 to 7 sentences to provide more \textit{context} per document, as larger chunks tend to yield fewer but more generalized topics. 
The BERTopic pipeline was tuned/adapted accordingly: UMAP parameters were adjusted to promote broader clustering (\texttt{n\_neighbors} = 16, \texttt{min\_dist} = 0.2, \texttt{n\_components} = 4), and HDBSCAN settings were modified to favor stable clusters (\texttt{min\_cluster\_size} = 11, cluster selection = \texttt{eom}). These changes bias the model toward extracting higher-level themes that recur across interviews. 
We also revised the LLM-prompt labeling process to suit global themes. Instead of using representative documents, the new LLM prompt focuses \textit{exclusively on keywords} for each topic:
\begin{quote}
``You are an AI that labels discussion topics, from a collection of cancer interviews, for a software that allows doctors to browse through medical files without the need to read them from start to finish. Given the following keywords, provide a clear and specific topic label, with enough context to be interpretable, focusing on the keyword list. Be general. Keep it short. Only type the topic label and nothing else."
\end{quote}
This approach ensures more objective and consistent labels for overarching themes. Additionally, we expanded the stop word list to reduce noise, though translation errors and linked words across interviews made perfect filtering impractical without manual review.

\subsubsection{Model Output: Topic Analysis}
Using the above configuration and parameters, we fitted the topic model on the entire corpus of 13 interviews. The resulting output (Table \ref{tab:globalmodeloutput}) reveals overarching themes that span all interviews, offering a broad view of the cancer treatment experience. 
Unlike granular per-interview models, which capture individual nuances, the global model identifies recurrent concerns and \textit{shared points across patients}, enabling detection of systemic issues and common emotional or physical pain points. 
For example, \textbf{Topic 0} includes keywords such as ``oxycontin'', ``medications'', and ``nausea'', highlighting widespread struggles with medication side effects. \textbf{Topic 2} reflects how patients recall surgical interventions, specifically keyhole surgery, while \textbf{Topic 7} clusters references to timelines and appointments, suggesting opportunities for visual treatment timelines. Similarly, \textbf{Topic 4} (Sleep Patterns and Nighttime Activities) emphasizes recurring sleep-related discussions, indicating that sleep is a significant aspect of the cancer journey and warrants deeper analysis to uncover common issues affecting rest and well-being.

Other topics capture \textit{institution}-specific and emotional insights. \textbf{Topic 9} focuses on ``Support and Resources for Cancer Care at Erasmus MC'', referencing amenities like food, lighting, and buildings. These are elements that, while non-clinical, influence patient experience and could inform facility improvements. 
\textit{Emotional} responses also emerge prominently: \textbf{Topic 8}, labeled ``Coping with Treatment Setbacks and Emotional Reactions'', groups keywords related to coping mechanisms, underscoring the need for mental health integration in oncology care. Finally, \textbf{Topic 12} (Navigating Treatment Decisions with Specialist \textit{Nurses}) highlights the critical role of nurses in guiding patients through complex choices, providing both clinical clarity and emotional reassurance. Overall, the globally tuned model captures high-level patterns across medical procedures, emotional resilience, logistics, and support systems, complementing per-interview analyses. While minor issues like repetitive keywords (e.g., ``day day'' in \textbf{Topic 0}) persist, they are less pronounced, confirming the effectiveness of this broader modeling approach.

\begin{table}[t]
\centering
\caption{BERT-topic Global Analysis using BioClinicalBERT embedding on All 13 Interviews}
\resizebox{\textwidth}{!}{
\scriptsize
\begin{tabular}{|l|p{6cm}|p{8cm}|}
\hline
\textbf{Topic ID} & \textbf{Topic Label} & \textbf{Top 15 Keywords} \\ \hline
0  & Medication Management and Symptom Relief in Cancer Care & knee, day day, pills, oxycodone, medications, diarrhea, medication, symptoms, times day, went doctor, nausea, stomach, prescribed, consultantdoctor, ones \\ \hline
1  & Impact of Chemotherapy on Patient Experience and Expectations & chemo, chemotherapy, intense, effects, tomorrow, paper, oncologist, does thats, start chemo, took long, oncologist oncologist, door, meeting, cells, drive \\ \hline
2  & Placement of Portacath via Keyhole Surgery & portacath, placed, arm, keyhole surgery, keyhole, anesthesia, surgery, portacath portacath, probe, puncture, puts, run, shower, sedated, poked \\ \hline
3  & Nutrition and Dietary Habits in Cancer Care & cook, drinking, eating, sandwich, food drink, taste, eat, eating drinking, food, dietician, weight, eaten, brother, soup, fat \\ \hline
4  & Sleep Patterns and Nighttime Activities & sleep, downstairs, bed, couch, awake, lie, bathroom, watch, single, wash, groceries, rest rest, outside, upstairs, cup \\ \hline
5  & Family Support and Life Impact in Cancer Journeys & son, sister, joint, mother, project, children, twice, life, lives, times time, live, kind thing, older, child, large \\ \hline
6  & Diagnostic Imaging and Tests for Abdominal Conditions & bowel, ultrasound, mri, stomach, examination, appendix, tests, ct, biopsy, admission, ct scan, medium, pain clinic, scan hospital, taken \\ \hline
7  & Radiation Therapy Treatment Experiences and Side Effects & radiotherapist, radiation, courses, treatments, poked, december, october, thats possible, abdominal pain, abdominal, operate, markers, september, november, placed \\ \hline
8  & Coping with Treatment Setbacks and Emotional Reactions & failed, plan, weeks later, wall, reactions, success, calmly, tremendous, dirty, alive, face, march, nerves, cells, tried \\ \hline
9  & Support and Resources for Cancer Care at Erasmus MC Hospital & euros, erasmus, erasmus mc, mc, places, hospitals, hospital erasmus, light, food drink, support, hair, approach, lot people, possibly, building \\ \hline
10 & Patient Experience with Doctor Appointments and Risk Assessment & risk, doctor hospital, date, wonder, gosh, appointment doctor, forget, data, rotterdam, touch, tomorrow, space, ended, wife, march \\ \hline
11 & Patient Experience with Medical Equipment and Care Delays & pump, ticket, broken, waited, burden, air, walked, nurses, hours, does work, minutes, outpatient, decisions, early, nursing \\ \hline
12 & Navigating Treatment Decisions with Specialist Nurses & decisions, treatment process, experiences, trajectory, open, calls, important decision, advise, negative, specialist nurse, shes, cries, real, super, metastatic \\ \hline
13 & CyberKnife Treatment Program in Rotterdam & rotterdam, program, cab, stone, cyberknife, quarter past, button, file, quarter, puncture, push, examination, family doctor, liver, record \\ \hline
14 & Coordination and Communication in Cancer Care Management & responsible, team, secretary, order, number, personal, doctor come, creon, surgeon, clear, scary annoying, short, conversations, turn, knows \\ \hline
\end{tabular}
}
\label{tab:globalmodeloutput}
\end{table}

\subsubsection{Topic Distribution in the whole corpus}

\begin{figure}[t]
    \centering
    \includegraphics[width=1\textwidth, height=0.50\textheight]{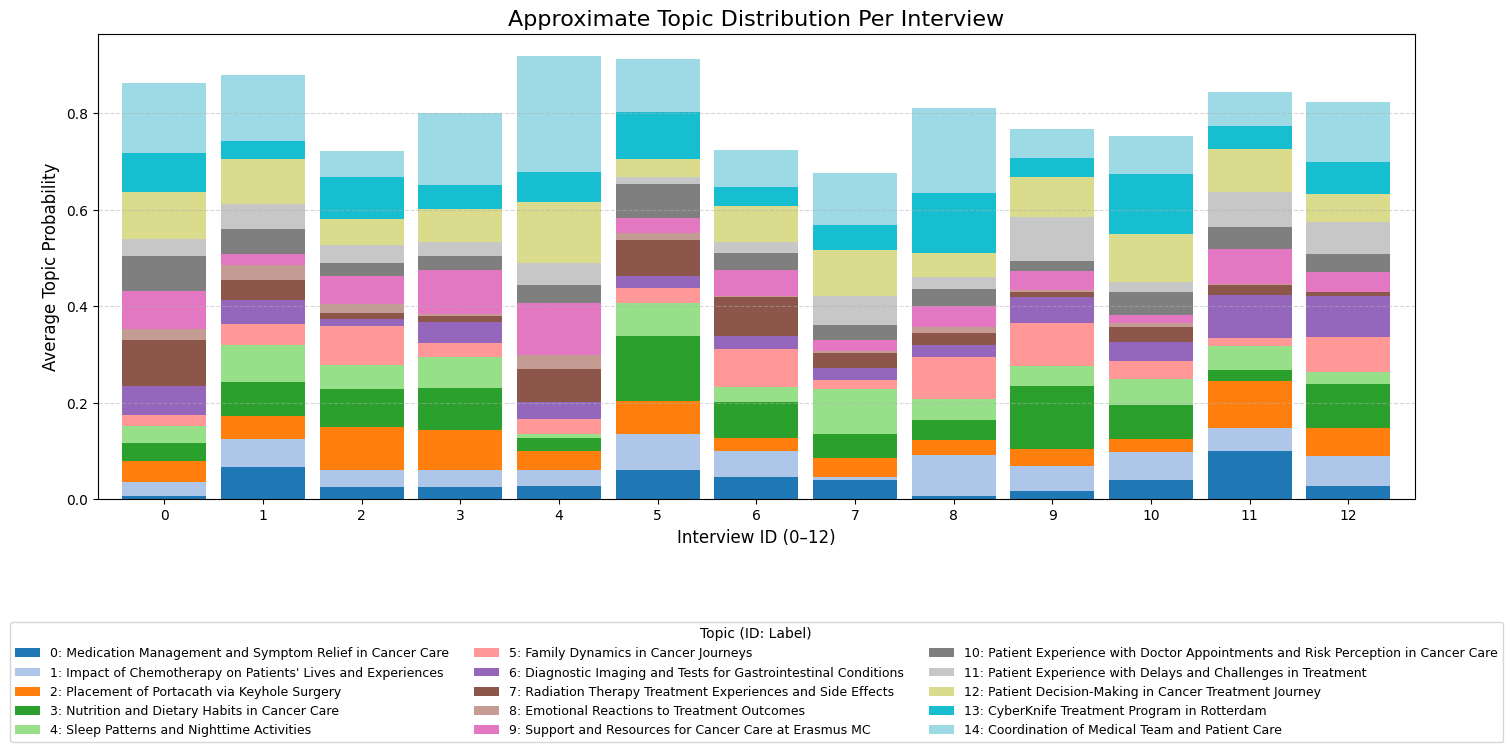}
    \caption{Approximate distribution across all 13 interviews, where each color represents a topic, and each bar represents one of the 13 interviews.}
    \label{fig:approxdis}
\end{figure}

Figure~\ref{fig:approxdis} shows the distribution of topics over interviews. Rather than assigning a single topic to each chunk, the distribution is based on the soft output of the BERTopic model.\footnote{This functionality is provided directly by BERTopic through its \texttt{approximate\_distribution()} method, which estimates the topic probabilities for each chunk without requiring re-fitting the model.} This allows for the possibility that a chunk touches on multiple themes to varying degrees. By averaging these probability distributions across all chunks in a given interview, an interpretable vector is produced that captures how strongly each topic is present throughout the interview as a whole. This approach is useful for capturing \textbf{thematic overlap} and uncertainty that may be missed with hard topic assignments. Table~\ref{tab:approxmost} lists the 5 highest probability topics across all 13 interviews.

The results indicate that \textbf{Topic 14} (Coordination and Communication in Cancer Care Management) appears predominantly across several interviews (e.g. Interviews 0, 1, 3, 4, 8), even in those where it was not one \revision{of the top three} (most notably, Interview 3). This suggests that while it may not have been the main focus in any one interview, it is a persistent theme that underlies many conversations, with the coordination of the medical team likely being a common underlying conversational topic among all patients. Another shift of this sort occurs with \textbf{Topics 12} (Navigating Treatment Decisions with Specialist Nurses) \textbf{and 13} (CyberKnife Treatment Program in Rotterdam), which are rarely top topics by prevalence, but frequently show up in the approximate distribution. This implies these themes may appear in shorter or more subtle forms, mentioned briefly, but across a \textit{wide range of patients}. Meanwhile, the sharply focused topics (such as \textbf{Topics 0 or 2}) remain central only to select individuals.

\begin{table}[t]
\centering
\caption{Most Occurring Topics Overall (Approx. Distribution)}
\begin{tabular}{|l|l|l|}
\hline
Topic ID & Mean Avg. Probability & Topic Label \\
\hline
14 & 0.118 & Coordination of Medical Team and Patient Care \\ \hline
12 & 0.079 & Patient Decision-Making in Cancer Treatment Journey\\ \hline
3  & 0.071 & Nutrition and Dietary Habits in Cancer Care \\ \hline
13 & 0.070 & CyberKnife Treatment Program in Rotterdam\\ \hline
2  & 0.053 & Placement of Portacath via Keyhole Surgery\\
\hline
\end{tabular}
\label{tab:approxmost}
\end{table}

\section{Discussion}
\label{sec:discussion}
The experiments conducted in this study provide insight into the capabilities of BERTopic and Top2Vec for extracting clinically relevant themes from cancer patient interviews. Overall, BERTopic -- particularly when configured with clinically oriented embeddings and sentence-based chunking -- demonstrated a stronger ability to capture nuanced topics. These findings suggest that topic modeling could serve as the basis for clinical feedback tools, enabling medical staff to navigate lengthy, unstructured patient documents efficiently and focus on \textit{patient-centered} care. 
Comparing BERTopic and Top2Vec revealed clear differences: while both produced coherent topics, BERTopic consistently generated more precise and less overlapping themes, which is critical in medical contexts, while Top2Vec generated more descriptive key terms. The ability to leverage domain-specific embeddings such as BioClinicalBERT further improved performance, confirming that embedding selection plays a key role in topic coherence and relevance.

We would like to point out a few limitations. First, topic modeling proved highly sensitive to chunking strategies: smaller chunks increased granularity but fragmented narratives, while larger chunks preserved context but reduced specificity. This trade-off between granularity and context preservation proves to be a recurring theme in the experimental setup for both BERTopic and Top2Vec. Second, while substantial effort was made to optimize the processing of the data, it is possible that alternative approaches or configurations, particularly in areas such as preprocessing, embedding selection, or clustering strategies, could have yielded improved or different results. Third, our evaluation survey was small-scale and lacked clinical experts, meaning that judgments were based on interpretability rather than professional applicability. Dataset characteristics introduced further challenges: interviews were originally in Dutch and translated into English, which may have altered meaning.  These issues underscore broader resource limitations for non-English clinical NLP. Despite these challenges, the study demonstrates the potential of topic modeling as a foundation for clinical support tools that streamline workflows and amplify patient voices.

\paragraph{\revision{Practical Deployment and Generalizability}}

\revision{While our experiments are conducted on a controlled dataset, the proposed method is designed with scalability and adaptability in mind. In real-world applications, patient consultation datasets are typically larger and more heterogeneous, including variations in transcription quality (potentially using automated transcription methods), sub-domain coverage, and input formats.
First, the method can be extended to larger datasets without fundamental architectural changes. The core components (e.g., BERTtopic and Top2vec keyword list extraction and LLM-based topic labeling) scale linearly with the size of the data, and can be efficiently trained using standard batching and parallelization techniques commonly used in modern machine learning frameworks.
Second, the approach is robust to heterogeneous data. In practice, domain variation can be addressed through standard techniques such as domain adaptation, fine-tuning on target-specific data, or incorporating additional contextual features. The modular design of our method allows it to integrate such extensions without requiring substantial redesign. 
Finally, in a real-world deployment scenario, the method could be integrated into existing pipelines as a decision-support component. For example, it could be used to assist decision makers to gain insights of their patient treatment procedures.
This would allow practitioners to benefit from the model's predictions while maintaining human oversight where necessary. For a setting like that, a user interface is recommended that allows the user to link back the identified topics to the corresponding text chunks in the topic cluster of BERTopic.
Evaluating the system on larger and more diverse datasets is an important direction for future work, and we plan to explore this in subsequent studies.}

\section{Conclusions and Future Work}

Overall, this study highlights the value of embedding-based topic modeling for clinical applications on patient care. By automating the extraction of meaningful themes from patient narratives, tools like BERTopic could enhance workflow efficiency, awareness of patient concerns, and strengthen patient-doctor communication. 

\textbf{In response to the first research question}, \revision{we found that, while both techniques produce fairly coherent and easily interpretable topics, BERTopic, delivers more refined and focused results, which better represent the patients' experiences and concerns expressed during the interviews. The model effectiveness can be further improved} when paired with an embedding model pretrained on large amounts of clinical data, such as BioMedicalBERT, and a sentence-based chunking strategy. This makes BERTopic more promising for practical use in clinical environments, as opposed to Top2Vec, given the specific experimental setup and dataset. 

\textbf{As for the second research question}, \revision{the results show that current neural topic modeling techniques, namely Top2Vec and BERTopic, can extract a variety of relevant themes from patient interviews. These include emotional experiences, treatment details, personal struggles, and reflections on the treatment processes.} 

The extracted topics suggest several ways in which topic modeling could support and improve healthcare processes. Most importantly, they could help clinicians identify and understand key moments in a patient's narrative without having to read every transcript manually. This could save time, reduce the workload of clinical staff, and give more visibility to the patient's voice, especially in cases where emotional or psychological concerns might otherwise be overlooked. Although not empirically tested in this study, this type of automated workflow for analyzing patient documents could also help reduce the risk of overlooking important details in a patient's history. By eliminating the need for clinicians to read through lengthy transcripts manually, the system may lessen the chance of missing key information due to time constraints or fatigue. Although this study did not involve a clinical trial or professional assistance, the structure of the output, combined with feedback from the conducted small-scale human evaluation, shows clear potential for integrating topic modeling into tools that support patient-centered care. Moreover, the global analysis of all 13 interviews reveals recurring themes discussed by all patients, which could potentially assist in identifying patterns in cancer patient treatment journeys in order to mitigate common issues.

While this work remains a proof of concept, it lays the foundation for future research focused on multilingual capabilities, contextual models, 
improved evaluation frameworks, and integration into clinical decision-support systems. Such advancements could help transform narrative data into actionable insights, placing patient voices at the center of healthcare delivery.

For future work, it will be important to test the system with clinical experts to better understand how the generated topics can be applied/integrated in a real-world healthcare setting. 
Beyond that, moving past translated text is an important next step. This includes not only exploring Dutch-language embedding models suitable for processing the original interviews, but also gathering new datasets in English and other languages. Doing so would serve the purpose of supporting the development and training of more robust multilingual, domain-specific embedding models, but it would also allow this approach to be tested on native language data to see how it performs without the distortions that come with translation. 
Additionally, future experiments could explore more adaptive or dynamic chunking strategies, which might better balance granularity and contextual coherence, especially across interviews of different lengths and structures. 
Lastly, the global dataset analysis could be expanded to track how the identified global themes evolve across different patient populations over time (\textit{temporal} dimension), and offer solutions to help clinical staff solve or minimize common complaints from cancer patients.






\section*{\revision{Ethical considerations}}

\revision{Not all patient data can be used for automated analysis like topic modeling. The data used in this paper was provided to us by the 4D PICTURE project consortium. The implicit assumption is that topics extracted from a small number of patient interviews are potentially relevant for cancer patient care in the broad sense. We do not intend to extrapolate the topic modeling results to individual patient care, but use it to inform policies on patient communication: which aspects of living with cancer are important to patients in which stage of their disease.}

\section*{Acknowledgments}
We thank Ida Korfage and Sheila Payne for the valuable comments on the Abstract version of this work (presented at CLIN2025).
We thank Anne Kuld-Nielsen for valuable feedback and encouragement from reading the manuscript.

This work has been funded by the European Union under Horizon Europe Work Programme 101057332. Views and opinions expressed are however those of the author(s) only and do not necessarily reflect those of the European Union or the European Health and Digital Executive Agency (HaDEA). Neither the European Union nor the granting authority can be held responsible for them.
The UK team members are funded under the Innovate UK Horizon Europe Guarantee Programme, UKRI Reference Number: 10041120.

\revision{The data originates from the project ``Design for Shared Decision Making: to support a team and enhance service Resilience'' funded by KWF under number UL-12072.
}

\bibliographystyle{clin} 
\bibliography{bibliography}  

\appendix
\section{Volunteer Survey Questionnaire}
\label{ref:appendixa}
This section contains the volunteer survey questionnaire, analyzed in Section \ref{sec:modelanalysis}, which we used in order to evaluate the outputs of the two chosen topic modeling techniques. The volunteers were presented with interview I0, along with the following survey to complete:

\bigskip

\begin{quote}
    \textit{Context: This survey is part of a research project investigating the use of automated topic modeling techniques on cancer patient interview transcripts. The broader goal of this thesis is to explore how topic modeling can help clinical staff quickly extract relevant insights, such as emotional responses, symptoms, experiences, etc., from lengthy patient narratives without having to read entire files. You will be presented with one anonymized patient interview and the resulting topic outputs from each model. Your feedback will help evaluate the clarity, relevance, and usefulness of the topics, contributing to an assessment of how well these models could support future clinical decision-making tools.\\
    \\
    On the next page, you will find the extracted topics and keywords generated by each model. Please read the interview first before proceeding with the questions.}
\end{quote}
\begin{enumerate}
  \item \textbf{Please rate each topic on a scale from 1 to 5, where:}
  \begin{itemize}
    \item 1 = Not coherent / Not useful
    \item 5 = Very coherent / Very useful
  \end{itemize}
  For each rating, please add a brief explanation if needed.\\
  \textit{Example: T1: 4 – Useful topic, but title could be more precise.\\
  T2: 1 - The topic makes no sense, this was never talked about during the interview.}

  \item \textbf{Overall, how well do the extracted topics represent the content of the interview you read?}\\
  (1 = Not at all, 5 = Very accurately)

  \item \textbf{How helpful/accurate were the keywords under each topic for understanding what the topic was about?}\\
  (1 = Not helpful, 5 = Very helpful)\\
  \textit{Example: T1: 4 - Useful keywords, but it has one irrelevant word in it: "the")}

  \item \textbf{Were there any important themes, ideas, or aspects of the interview that were missing from the extracted topics?}
  \begin{itemize}
    \item Yes
    \item No
  \end{itemize}
  If yes, please specify.

  \item \textbf{Do you have any additional feedback or suggestions about the topics or the overall experience?}
\end{enumerate}

Each participant received one survey document for each model, which had the same questions, with the only difference being the topic output. 

\section{\revision{Clinically-Oriented Embedding Model Outputs}}
\label{ref:appendixb}
\revision{The analysis of the three clinically-oriented embedding models, namely \textbf{BioClinicalBERT}, \textbf{ClinicalBERT}, and \textbf{MSR BiomedBERT}, previously known as \textbf{PubMedBert}, can be found in Section \ref{sec:furtherexp}. The settings for the final topic outputs are the following:}

\revision{\begin{itemize}
    \item \textbf{ClinicalBERT}: Baseline settings + 7-sentence chunking + \texttt{min\_df = 3} + \texttt{n\_components = 8}
    \item \textbf{BioClinicalBERT}: Baseline settings + 7-sentence chunking + \texttt{min\_df = 3}
    \item \textbf{MSR BiomedBERT}: Baseline settings + 6-sentence chunking + \texttt{min\_df = 3}
\end{itemize}}

\revision{The output for each model, which includes every topic label and top 15 keywords for each topic, can be found below for each of the embedding models. As with the rest of the outputs showcased throughout this research, the following outputs are also obtained using interview I0.}

\begin{table}[h]
\centering
\caption{Final Output with the ClinicalBERT-Tuned BERTopic Model (Interview I0)}
\resizebox{\textwidth}{!}{
\scriptsize
\begin{tabular}{|l|p{5cm}|p{8cm}|}
\hline
\textbf{Topic ID} & \textbf{Topic Label} & \textbf{Top 15 Keywords} \\ \hline
0  & Challenges in Patient-Doctor Conversations Within Limited Treatment Spaces and Options & conversation, room, started, probably, little bit, little, wait, cures, eye, let, june, 19, long time, things, called \\ \hline
1  & Patient Engagement in Cancer Care: Importance of Asking Questions and Seeking Information During Appointments & questions, person, especially, felt, question, head, making, ask, asked, appointments, guys, mind, pancreatic, talk, patient \\ \hline
2  & Challenges with FOLFIRINOX Treatment, Size Discrepancies, and Information Gaps in Patient Experience & size, folfirinox, example, eventually, 19, new, information, let, day, prick, people, happens, happy, rotterdam, happen \\ \hline
3  & Emotional Impact of Cancer Diagnosis and Treatment Experiences at Daniel den Hoed Hospital & head, lying, daniel den, hoed, den hoed, daniel, den, grumpy, understand, write, going happen, couple times, rest, throat, point \\ \hline
4  & Delays in Chemotherapy Start Due to Treatment Coordination and Malignant Diagnosis Concerns & poked, june, malignant, radiotherapist, long time, 21, fact, able, chemo, follow, ultrasound, appointment, certain, radiation treatments, asked \\ \hline
5  & Delay in Tumor Marker Evaluation and Persistent Pain Leading to Further Medical Intervention & pain, ultrasound, tumor marker, marker, days, week, tumor, possible, weeks later, wait, contact, eye, rest, right, malignant \\ \hline
6  & Challenges in Scheduling Appointments and Blood Puncture Procedures During Cancer Treatment & appointments, december, puncture room, puncture, talk, appointment, end, anymore, need, blood, september, results, hand, patient, day \\ \hline
7  & Patient Experience and Choices in Hospital Transitions and Care Interactions & notice, people, real, sweet, choice, quite true, does matter, secretary, showed, nice, results, allowed, patient, matter, exactly \\ \hline
8  & Challenges and Experiences with Port-a-Cath Insertion and Blood Draw Procedures & looks, markers, port cath, cath, port, markers placed, prick, size, poked, blood, times, placed, hand, explained, happen \\ \hline
9  & Conversations with Doctors and Hospital Visits in April Regarding Patient Care & april, doctor doctor, conversation, touch, place hospital, hospital, erasmus, doctor, surgeon, idea, read, space, new, went, away \\ \hline
10 & Rising Tumor Markers and Scan Results Impacting Treatment Decisions in December 2017 & december, year, anymore, tumor marker, marker, blood, cure, tumor, scan, possible, takes, clear, sat, gee, heard \\ \hline
11 & Challenges in Communication with Medical Staff and Coping with Cancer Journey & secretary, difficult, life, learned, doing, prepared, rotterdam, knows, mean, possible, nurses, life coach, coach, hold, pick \\ \hline
12 & Navigating Hope and Fear in Pancreatic Cancer Diagnosis and Treatment Experiences & hope, pancreatic cancer, pancreatic, beginning, information, pain, sit, lot, mean, cancer, look, possible, ask, bad, definitely \\ \hline
13 & Waiting Room Experiences and Communication About Test Results in Cancer Care & getting, sitting waiting, guys, showed, television, taken, waiting room, metastases, waiting, hour, true, scan, half, sitting, moment \\ \hline
14 & Importance of Direct Consultation with Expert Surgeon in Rotterdam for Cancer Treatment & friday, rotterdam, does matter, matter, called, sitting, appointment, surgeon, sure, make sure, saying, brought, examination, definitely, follow \\ \hline
\end{tabular}
}
\label{tab:clinicalbert_labels}
\end{table}
\begin{table}[h]
\centering
\caption{Final Output with the BioClinicalBERT-Tuned BERTopic Model (Interview I0)}
\resizebox{\textwidth}{!}{
\scriptsize
\begin{tabular}{|l|p{5cm}|p{8cm}|}
\hline
\textbf{Topic ID} & \textbf{Topic Label} & \textbf{Top 15 Keywords} \\ \hline
0  & Patient Experience and Information Gaps in Pancreatic Cancer Treatment Choices and Communication & questions, especially, choice, guys, quite, information, pancreatic cancer, saying, quite true, pancreatic, kind, eventually, real, hold, story \\ \hline
1  & Challenges and Experiences with Port-a-Cath Usage During Cancer Treatment & prick, maybe, story, pain, port, port cath, cath, mean, prepared, sit, cure, beginning, look, possible, blood \\ \hline
2  & Experience of Radiation Treatment: Challenges, Waiting Times, and Patient Comfort During Procedures & idea, lie, weird, hour, stop, sit, radiation, exciting, people, helped, fine, conversation, end, gee, certain point \\ \hline
3  & Experiencing Anxiety and Uncertainty During Throat Examination and Treatment Processes & throat, quickly, going happen, make, understand, rest, couple times, lying, lie, happens, information, tubes, happen, times, doing \\ \hline
4  & Frustrations with Treatment Plans and Follow-Up in Cancer Care Conversations & plan, little bit, bit, probably, takes, wait, poked, cures, look, june, television, radiation treatments, guys, stop, place \\ \hline
5  & Monitoring Tumor Markers and Blood Sampling Experiences Over Time in Cancer Care & prick, blood, year, cures, tumor marker, couple, waiting room, december, times, tumor, weeks, waiting, marker, scan, option \\ \hline
6  & Frustration with Appointment Changes and Desire for Consistency in Doctor Consultations & grumpy, appointment, called, does matter, different, doctor doctor, matter, make sure, sure, monday, surgeon, nurse, afternoon, researcher, end \\ \hline
7  & Early Detection and Monitoring of Tumors Through Ultrasound and Tumor Markers & pain, heard, ultrasound, tumor marker, taken, saw, tumor, marker, beginning, true, hospital place, sitting waiting, gets, scan, real \\ \hline
8  & Patient Reflections on Communication and Information During Cancer Diagnosis and Treatment Journey & sorry, especially, remember, information, hear, bed, pretty, throat, lying, minutes, later, question, tubes, examination, room \\ \hline
9  & Malignant Biopsy Experiences and Challenges in the Puncture Room at Daniel den Hoed & malignant, puncture room, puncture, den hoed, daniel den, hoed, den, daniel, person, ultrasound, blood, results, saw, hand, came \\ \hline
10 & Discussion on Pancreatic Cancer Diagnosis and Treatment Experiences in Rotterdam, April Timeline, and Communication & hear, getting, rotterdam, guys, size, long time, obviously, april, pancreatic cancer, true, surgeon, metastases, pancreatic, remember, important \\ \hline
11 & Timeline of Cancer Treatment: Appointments, Chemo Start Dates, and Patient Experience & 21, september, chemo, june, walking, follow, end, took, town, appointment, radiation, asked, december, start, long time \\ \hline
12 & Understanding Port-a-Cath Placement and Tumor Markers in Cancer Treatment Context & looks, markers placed, placed, rotterdam, markers, cath, port, port cath, explained, exciting, possible, nurse, ask, tumor, happy \\ \hline
13 & Navigating Life After Cancer: Learning, Recovery, and Uncertainty in Personal Experiences & life, learned, weird, knows, doing, little bit, bit, pick, took long, hold, half, town, moment, goes, room \\ \hline
14 & Impact of Personal Connections and Delays on Patient Experience in Healthcare Settings & notice, people, sweet, showed, secretary, does matter, results, date, nice, matter, patient, does, stuff, feel, important \\ \hline
15 & Significant Conversations and Key Dates Related to Doctor Visits at the Hospital & april, date, conversation, place hospital, doctor, space, hospital, read, doctor doctor, called, erasmus, surgeon, good, away, new \\ \hline
16 & Key Moments and Experiences During Cancer Treatment Journey: Insights from Patient Perspectives & moments, wrong, times, let, happened, experienced, researcher, gee, happy, certainly, everybody, eventually, appointments, rest, couple times \\ \hline
\end{tabular}
}
\label{tab:bioclinicalbert_labels}
\end{table}

\begin{table}[h]
\centering
\caption{Final Output with the MSR BiomedBERT-Tuned BERTopic Model (Interview I0)}
\resizebox{\textwidth}{!}{
\scriptsize
\begin{tabular}{|l|p{5cm}|p{8cm}|}
\hline
\textbf{Topic ID} & \textbf{Topic Label} & \textbf{Top 15 Keywords} \\ \hline
0  & Delay in Chemotherapy Start Date and Emotional Impact on Patient Care & head, tubes, stuff, 21, june, date, saw, gone, chemo, examination, placed, puncture room, puncture, fact, lie \\ \hline
1  & Patient Experience and Information Seeking During Cancer Treatment Journey in Rotterdam & comes, stop, mind, prepared, rotterdam, mention, feel, showed, hold, later, pancreatic cancer, pancreatic, googled, bed, kind \\ \hline
2  & Sudden Rise in Tumor Marker Levels and Impact on Patient Care Decisions & september, year, tumor marker, suddenly, end, days, blood, tumor, cure, marker, weeks, scan, radiation, went, weeks later \\ \hline
3  & Patient Experience with Hospital Procedures and Communication Challenges in Rotterdam & mean, couple times, understand, lie, 19, does, read, rotterdam, allowed, speak, brought, nurses, times, sitting waiting, experienced \\ \hline
4  & Importance of Patient Questions and Understanding in Cancer Treatment Process and Port-a-Cath Use & markers placed, questions, important, placed, cath, port, port cath, process, treatment process, moments, markers, ask, kind, moment, looks \\ \hline
5  & Coordination of Medical Care: Conversations and Travel to Rotterdam for Pancreatic Cancer Treatment & doctor doctor, friday, rotterdam, conversation, pick, surgeon, googled, definitely, doctor, took, went, plan, april, idea, read \\ \hline
6  & Urgent Need for Follow-Up Appointments with Radiotherapist and Surgeon Amidst Treatment Changes & suddenly, radiotherapist, does matter, able, appointment, grumpy, matter, ultrasound, asked, surgeon, talk, room, sitting waiting, right away, waiting room \\ \hline
7  & Importance of Patient Experience and Communication in Cancer Care at Hospitals & patient, important, place hospital, sweet, number, researcher, tubes, examination, malignant, doctor, understand, half hour, grumpy, long, den \\ \hline
8  & Discussion on Surgical Procedures, Patient Experiences, and Emotional Responses Related to Cancer Treatment & gee, exciting, 2017, puncture room, puncture, saying, surgery, clear, true, quite true, year, poked, 19, working, sorry \\ \hline
9  & Monitoring Eye Health and Treatment Progress in Cancer Care Discussions & eye, keeps, june, started, probably, radiotherapist, 19, making, cures, look, knows, maybe, looks, bit, need \\ \hline
10 & Experiences with Radiation Treatment at Daniel den Hoed for Pancreatic Cancer & den hoed, hoed, den, daniel, daniel den, idea, radiation, thought, possible, remember, scary, exactly, bed, option, lie \\ \hline
11 & Patient's Proactive Approach in Seeking Second Opinions and Treatment Options During Consultations & minutes, ask, look, fine, helped, list, option, definitely, getting, fact, asked, real, treatment process, process, folfirinox \\ \hline
12 & Challenges and Experiences with Blood Draws and Port-a-Cath During Cancer Treatments & poked, prick, half hour, blood, results, radiation treatments, waiting, treatments, port, cath, port cath, hour, times, cath port, feel \\ \hline
13 & Hope and Uncertainty Surrounding Pancreatic Cancer Diagnosis and Treatment Experiences & hope, cure, days, throat, maybe, bad, pancreatic cancer, pancreatic, does work, number, scary, gone, keeps, explained, couple times \\ \hline
14 & Challenges and Decisions in Cancer Treatment: Folfirinox, Metastases, and Surgical Options & folfirinox, metastases, wait, easy, getting, make, surgery, television, probably, way, start, sorry, cath, port, port cath \\ \hline
15 & Experiencing Cold, Unwelcoming Spaces and Lack of Choice in Medical Settings & exactly, real, quite true, moments, sitting, anymore, choice, space, felt, read, room, need, allowed, hospital, talk \\ \hline
16 & Issues with Tumor Marker Monitoring and Communication of Treatment Plans During Appointments & plan, tumor marker, came, tumor, new, marker, make sure, sure, learned, come, clear, television, appointment, went, certain \\ \hline
\end{tabular}
}
\label{tab:msrbiomedbert_labels}
\end{table}


\end{document}